\documentclass[letterpaper]{article} 
\usepackage{aaai2026}  
\usepackage{times}  
\usepackage{helvet}  
\usepackage{courier}  
\usepackage[hyphens]{url}  
\usepackage{graphicx} 
\usepackage{natbib}  
\usepackage{caption} 
\usepackage{algorithm}
\usepackage{algorithmic}
\usepackage{multirow}
\usepackage{hyperref}
\usepackage{newfloat}
\usepackage{listings}
\DeclareCaptionStyle{ruled}{labelfont=normalfont,labelsep=colon,strut=off} 
\floatstyle{ruled}
\newfloat{listing}{tb}{lst}{}
\floatname{listing}{Listing}
\title{High-Quality Full-Head 3D Avatar Generation from Any Single Portrait Image}
\newcommand{\sharedfootnote}{\footnote{\textit{Corresponding authors.}}}
\author{
   Yujie Gao\textsuperscript{\rm 1}, Chencheng Wang\textsuperscript{\rm 1}, Xianbing Sun\textsuperscript{\rm 1}, Jiahui Zhan\textsuperscript{\rm 1}, Wentao Wang\textsuperscript{\rm 2}, Yiyi Zhang\textsuperscript{\rm 1}, Haohua Zhao\textsuperscript{\rm 1}, Liqing Zhang\textsuperscript{\rm 1}\sharedfootnote, Jianfu Zhang\textsuperscript{\rm 1}\footnotemark[\value{footnote}]
}
\affiliations{
    \textsuperscript{\rm 1}Shanghai Jiao Tong University,\\
    \textsuperscript{\rm 2}Shanghai AI Laboratory\\
    \{GYjgyj123, nyte\_plus, fufengsjtu, jiahuizhan, yi95yi, haoh.zhao, zhang-lq, c.sis\}@sjtu.edu.cn; \\
    wangwentao@pjlab.org.cn;

}

\usepackage{bibentry}

\begin{document}

\maketitle
\begin{figure*}[h!]
    \centering
    \includegraphics[width=0.88\textwidth]{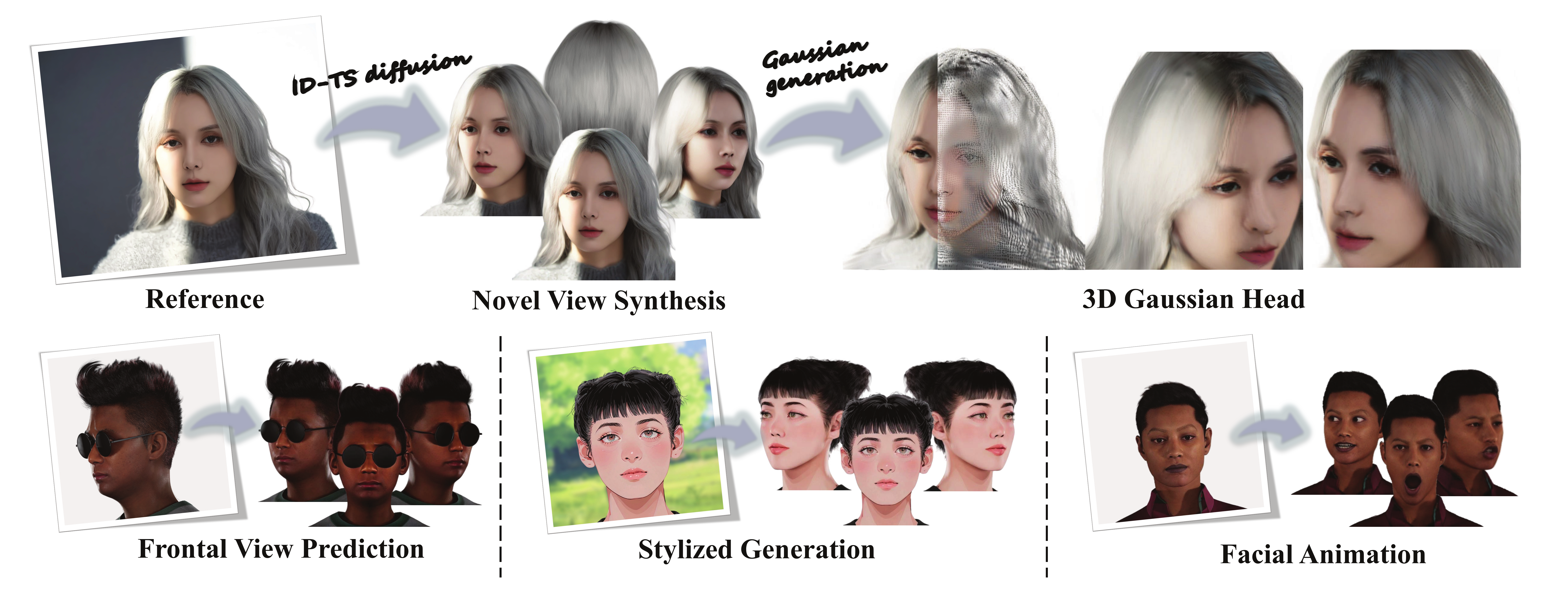} 
    \caption{Given a single head image, our method can generate multi-view images with rich facial details. And the high-fidelity 3D Gaussian head of the identity is then reconstructed using the multi-view images. Our method also generalizes well to multiple downstream tasks, such as  frontal face prediction, stylized head generation, and facial expression animation.}
    \label{fig:teaser}
\end{figure*}

\begin{abstract}
In this work, we introduce a novel high-fidelity full-head 3D avatar generation method from a single image, regardless of perspective, style, expression, or accessories. Prior works often fail to preserve consistent head geometry and facial details, primarily due to their limited capacity in modeling fine-grained facial textures and maintaining identity information. To address these challenges, we construct \textbf{a new high-quality dataset} containing 227 sequences of digital human portraits captured from 96 different perspectives, totaling 21,792 frames, featuring high-quality facial texture details. To further improve performance, we propose a novel multi-view diffusion model named \textbf{ID-TS diffusion model}, which integrates identity and expression information into the two-stage multi-view diffusion process. The low-resolution stage ensures structural consistency of heads across multiple views, while the high-resolution stage preserves facial detail fidelity and coherence. Finally, we propose \textbf{an enhanced feed-forward Gaussian avatar reconstruction method} that optimizes the network on multi-view images of each single subject, significantly improving 3D facial texture details. Extensive experiments demonstrate that our method achieves robust performance across challenging scenarios, while showcasing broad applicability across numerous downstream tasks. Our project page is available at \href{https://biggaoga.github.io/HQ-Head-Generation/}{https://biggaoga.github.io/HQ-Head-Generation/}.
\end{abstract}


\section{Introduction}
Synthesizing full-head 3D models with arbitrary expressions has widespread applications in  game design, AR/VR, video conferencing,  \textit{etc}. 
Traditional approaches rely on statistical 3D face models~\cite{blanz2023morphable,deca,wang20243d} to generate high-fidelity 3D heads.  
However, they are typically limited to the facial region and often fail to reconstruct the full head, missing important features such as hair, glasses, and accessories. 
Other works~\cite{portrait4d,gpavatar,gagavatar,real3d,cap4d} have proposed methods for reconstructing 4D dynamic head models from single portrait images, achieving promising results. Nonetheless, these approaches share the same limitation that their focus remains primarily on the facial area, resulting in incomplete full-head reconstructions. This significantly constrains their applicability in downstream tasks.
 
In recent years, numerous single-image-to-3D reconstruction methods~\cite{long2024wonder3d,liu2023zero,shi2023zero123++,wang2023imagedream, instantmesh} based on multi-view diffusion models have achieved remarkable success, which inspires us to explore their potential for 3D human head generation.
While these models are trained on large-scale 3D datasets to generate multi-view images for 3D object reconstruction, they often lack sufficient geometric and textural consistency when the input is a human face. 
This discrepancy becomes particularly problematic because human faces require not only geometric consistency but also the preservation of subtle features such as expression, identity, and accessories.
Upon reviewing existing image-to-3D methodologies, the following primary limitations can be identified:
1) Most image-to-3D diffusion models are trained on large-scale 3D datasets~\cite{deitke2023objaverse,deitke2024objaverse}, which often lack high-quality head data and suffer from a limited number of meshes, resulting in poor quality reconstructions. Besides, existing facial datasets \cite{ava-256,renderme360, nersemble,cafca} solely focus on frontal facial regions while neglecting critical side and rear head perspectives for full-head reconstruction.
2) Existing image-to-3D diffusion models still struggle to maintain identity and expression consistency for human head generation. This is particularly challenging because human faces require much higher consistency compared to other objects.
3) Prior works struggle to ensure consistent facial feature details across multi-view portrait images, \textit{e.g.}, the human eyes may remain highly sensitive to subtle discrepancies among different views.
4) Previous multi-view diffusion models employed suboptimal 3D reconstruction approaches that failed to preserve high-fidelity texture details in human head modeling.

To address these challenges, we enhance the existing image-to-3D methodology with several key innovations.
First, we construct a \textbf{3D digital human head dataset} with diverse attributes (\textit{e.g.}, hairstyle, skin color, age, gender, accessory, expression, \textit{etc}.) to fine-tune the multi-view diffusion model and 3D reconstruction model. 
This dataset is constructed using a high-quality digital human head engine, offering rich facial texture details and enabling the simulation of complex texture variations during facial expression changes. Equally crucial is our ability to freely configure camera parameters, which facilitates the acquisition of appropriate multi-view images for training purposes.
Second, we propose an \textbf{identity-guided two-stage diffusion model}, called ID-TS diffusion model, which is specifically designed for high-fidelity human head generation.
To address the limitation of existing methods in preserving facial identity and expression information, we propose the identity-aware diffusion framework that incorporates ID embeddings to guide the generation process, significantly enhancing the model's capability in identity preservation.
Toward consistent and high-fidelity facial details, we adopt a two-stage diffusion process: the low-resolution stage addresses shape consistency across multiple viewpoints of the head, while the high-resolution stage refines facial details by integrating outputs from the low-resolution stage, thereby further improving the coherence of facial features.
Finally, we employ an\textbf{ enhanced feed-forward Gaussian avatar reconstruction method} for high-quality 3D human head reconstruction, while jointly optimizing the feed-forward network. Specifically, in order to enhance the detail preservation and spatial coherence of 3D Gaussian heads, we fine-tune the network with our dataset and optimize each subject with the 16-frame multi-view outputs from ID-TS diffusion model.
Experimental results demonstrate that our method can generate vivid 3D human heads, even under challenging conditions such as profile-view inputs, stylized head models, and complex facial expressions.

In summary, the contributions of our work are as follows: 
\begin{itemize}
    \item We propose a high-quality human head dataset consisting of 227 sequences of digital human portraits captured from 96 different perspectives, totaling 21,792 frames.  
    \item We introduce an identity-guided two-stage diffusion model (ID-TS diffusion model), which significantly improves multi-view identity consistency and expression preservation while achieving finer facial detail synthesis.
    \item We present an enhanced feed-forward Gaussian avatar reconstruction method that significantly improves the fidelity of 3D human head details.
\end{itemize}

\begin{figure*}[t]
    \centering
    \includegraphics[width=0.88\linewidth]{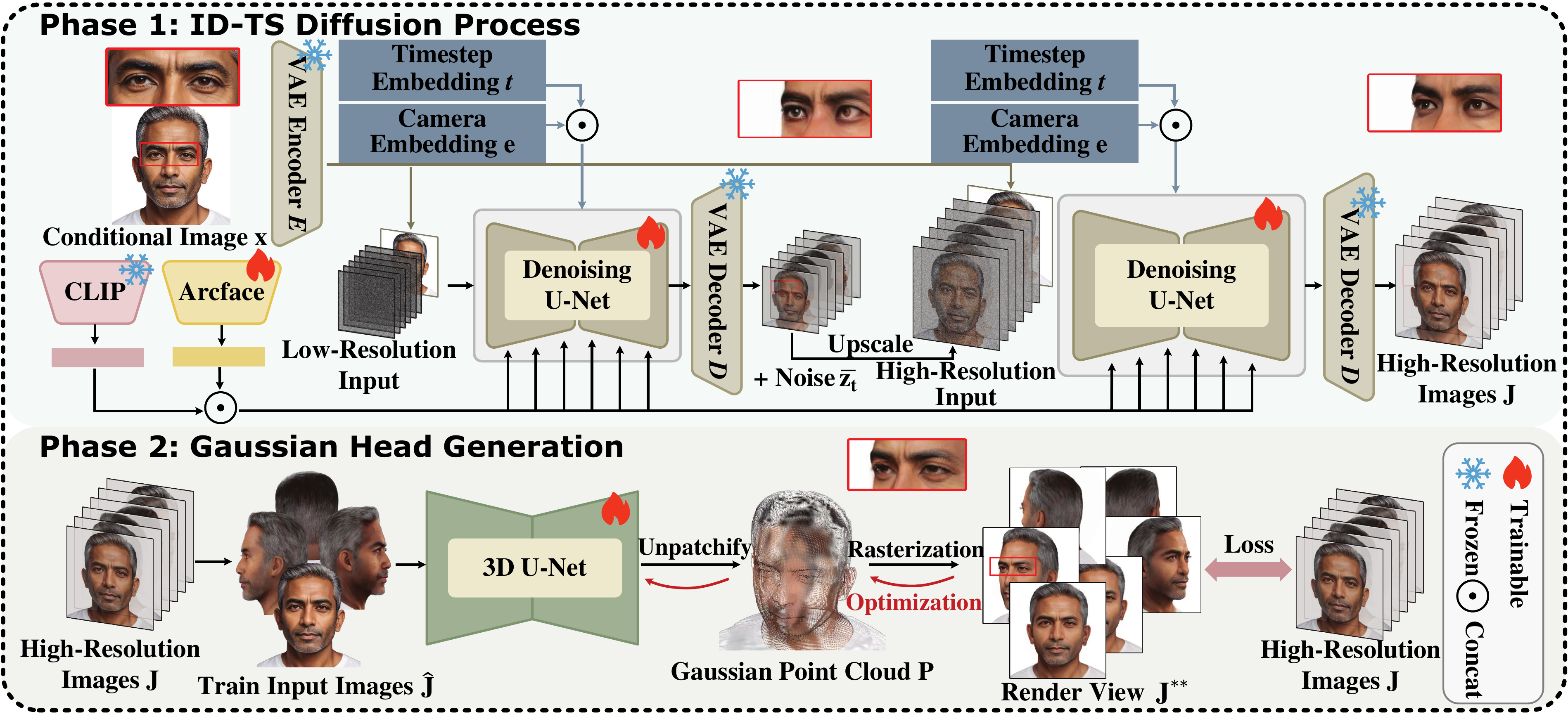}
    \caption{Overview of our inference pipeline. \textbf{Phase 1}: In the low-resolution stage, we embed camera pose \textbf{e} and noise step \textbf{t} via positional encoding, concatenate them, and feed them into the U-Net’s residual blocks. The conditional image \textbf{x} is encoded into the latent space of the VAE encoder\textbf{ E}, concatenated with noise, and processed alongside CLIP and ArcFace embeddings via cross-attention in the transformer block, generating multi-view images with accurate head shapes and coarse facial texture. In the high-resolution stage, we upsample the previous outputs, do element-wise addition with latent noise, and denoise them, outputting comprises multi-view images with high-fidelity texture details. \textbf{Phase 2}: With front/left/back/right images as inputs and remaining frames as supervision for 3D U-Net optimization, we finally yield a high-quality Gaussian head \textbf{P}.}
    \label{fig:pipeline}
\end{figure*}

\section{Related Works}

\noindent\textbf{3D Full-Head Generation:}
3D reconstruction of human head from a single image has been a hot topic for an extended period, resulting in numerous high-quality contributions to the field. PanoHead~\cite{panohead} builds upon the EG3D~\cite{chan2022efficient} and achieves full head synthesis based on GAN~\cite{goodfellow2014generative}. However, since its triplane representation is implicit, the expressive capability of their model is somewhat limited. ID-Sculpt~\cite{portrait3d}  uses ID-aware guidance for 3D head reconstruction from a single image and achieves relatively favorable results; but, its reliance on the head shape prior provided by PanoHead may lead to potential shape artifacts.  Rodin~\cite{rodin} and RodinHD~\cite{rodinhd} incorporate a combination of the diffusion model and the triplane representation. However, their methods fail to generate fine details and exhibit lower similarity with the input image. FaceLift ~\cite{facelift} conducts training on multi-view diffusion within a large-scale avatar dataset, aiming to generate high-fidelity 3D heads. However, its input images are restricted to the frontal views of identities, which poses a constraint on the data diversity and potential application scenarios. 

\noindent\textbf{Multi-View Diffusion Models:}
Multi-View diffusion builds upon text-to-image diffusion \cite{sd}, leveraging the strong 2D priors of diffusion models and achieving significant advancements through training on large-scale 3D datasets. Most methods focus their innovations on maintaining multi-view consistency. For instance, Zero123++~\cite{shi2023zero123++} fixes the camera viewpoint and employs a linear noise addition strategy, Wonder3D~\cite{long2024wonder3d} enhances consistency by controlling the generation of multi-view normal maps and RGB images, while ImageDream~\cite{wang2023imagedream} aligns text-to-3D diffusion model~\cite{shi2023mvdream} capabilities to ensure consistency in image-to-3D synthesis. However, their performance in multi-view facial synthesis is suboptimal, and the number of generated views is highly limited, which constrains the effectiveness of downstream 3D facial reconstruction tasks.

\noindent\textbf{3D Reconstruction:}
Reconstruction algorithms have made significant progress since the advent of NeRF~\cite{nerf}, which implicitly encodes a 3D scene into a neural network and renders it using volumetric functions. And there are also many methods~\cite{dreamfusion, wang2023imagedream} that apply NeRF for Score Distillation Sampling (SDS).
Other works~\cite{lrm} apply triplane for 3D scene encoding and utilize MLP to decode 2D images. However, the implicit representation restricts their applications, especially in object reconstruction. 
Recently, 3D Gaussian Splatting~\cite{3dgs} has demonstrated promising reconstruction results with the representation of Gaussian point cloud. This approach can be utilized for modeling dynamic scenes or objects, and its rendering speed is also fast. Nowadays, feed-forward Gaussian generation methods~\cite{lgm, gslrm} further improved multi-view consistency and surface smoothness in sparse view inputs, by integrating neural networks into Gaussian point clouds generation process and training in large-scale 3D dataset. However, the Gaussian point clouds generated through such a single forward pass often exhibit deficiencies in fine details, particularly for unseen objects.

\section{Methodology}
\subsection{Problem Formulation}\label{pf}
Given a single image $\textbf{x} \in \mathrm {R}^{3 \times H\times W } $ of an identity $\textbf{I}$ and the corresponding rough camera elevation $\textbf{e}$, we first encode this information into our ID-TS diffusion model. In the low-resolution stage, we generate multi-view images $\bar \textbf{J} \in \mathrm{R}^{N_s \times 3 \times h\times w}$, $N_s$ denotes image number, $h=H/\sigma$, $w=W/\sigma$, and $\sigma$ denoting the downsampling ratio. In the high-resolution stage, we upsample $\bar \textbf{J}$ and mix it with the random noise $\bar \textbf{z}_t \in \mathrm{R} ^{N_s \times 3 \times H\times W }$. The denoising process finally outputs multi-view images $\textbf{J} \in \mathrm{R} ^{N_s \times 3 \times H\times W }$ around the identity $\textbf{I}$. The camera positions of these output images are represented as  $\pi \in \mathrm{R} ^{N_s \times 2}=\{(\textbf{e}_i, \textbf{a}_i)\}_{i=1} ^{N_s}$, where $\textbf{e}_i$ corresponds to the elevation of each camera, equal to the input elevation  $\textbf{e}$, and $\textbf{a}_i$ represents the azimuth. After that, we optimize feed-forward Gaussian generation network to reconstruct the 3D head model using 4 of the multi-view images $\hat{\textbf{J}} \in \mathrm{R} ^{4 \times 3 \times H\times W }$ and the corresponding camera parameters $\hat{\pi}\in \mathrm{R} ^{4\times 2} = \{(\textbf{e}_j,\textbf{a}_j)\}_{j=1}^{4}$ as inputs and rest images as supervisory data. Finally, we get Gaussian point cloud $\textbf{P} \in \mathrm{R} ^{(4 \times H' \times W') \times 14}$, where $H'$ and $W'$ represent network output shape, the last dimension ``14'' includes 3-channel colors, 3-channel scale factors, 3-channel position coordinates, 4-channel rotation factors and 1-channel opacity factor. The total number of Gaussian points is $4 \times H' \times W'$.

 \subsection{Digital Human Head Dataset}\label{dhhd}
Existing large-scale 3D datasets for multi-view diffusion models suffer from several limitations: low-quality textures, insufficient mesh polygon counts, and a notable scarcity of 3D human head models. Furthermore, available public head datasets typically only provide limited viewpoint variations. For the purpose of getting high-fidelity 3D full-head models, we construct a digital human head dataset, which includes different genders, ages, races, \textit{etc}. Moreover, we provide high-fidelity texture details to facilitate network training. In total, we construct 227 sequences of 3D digital human portraits. \textbf{For detailed construction procedures, please refer to the supplementary material.}
 
\begin{figure}
    \centering
    \includegraphics[width=0.9\linewidth]{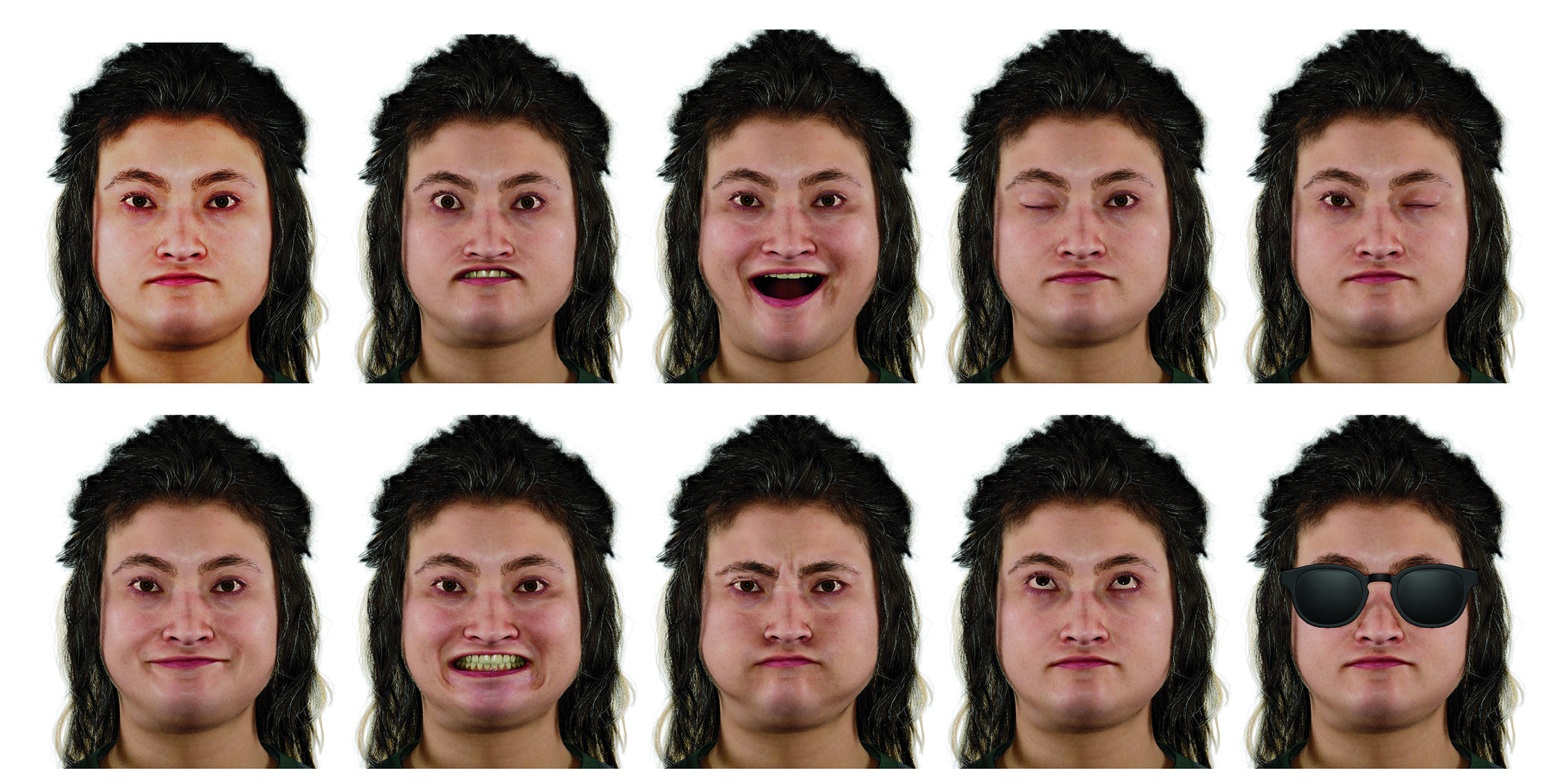}
    \caption{Different expressions and an accessory in 3D digital human models. Diverse facial expressions are sampled to enrich our dataset, along with an accessory to validate the fitting capability of our model.}
    \label{fig:exp}
\end{figure}

 To train our neural network, we sample multi-view images from the 3D models of digital human heads. Specifically, we render 1024×1024 resolution images from $N=100$ realistic human head models. These renderings cover  $N_e=6$  elevation angles of  $[-10^{\circ}, 0^{\circ}, 10^{\circ}, 20^{\circ}, 30^{\circ}, 40^{\circ}]$,  azimuth angles at $N_a=16$ positions around the head, resulting in $N_v=96$ unique viewpoints per head. Additionally, each avatar features $N_f=10$ appearances with $9$ different expressions and $1$ accessory, as shown in Fig.~\ref{fig:exp}.

\subsection{ID-TS Diffusion Model}\label{id-ts}
Multi-View image generation is a critical step in our method, providing the essential high-quality, identity-preserving, and appearance-consistent face images from different perspectives for 3D head reconstruction. 
We train our ID-TS diffusion model which takes one single image \textbf{x} as input and outputs spatially consistent multi-view images \textbf{J}. 
The entire process is illustrated in the higher segment of Fig.~\ref{fig:pipeline}.

\noindent\textbf{Architecture:}
ID-TS diffusion model consists of a two-stage diffusion model, and each diffusion model is initialized from Stable Video Diffusion~\cite{blattmann2023stable}. We adapt it into a multi-view generative framework, drawing inspiration from \cite{voleti2025sv3d,yang2024hi3d}.
For each diffusion model, the architecture contains VAE encoder \textbf{E}, decoder \textbf{D} and denoising U-Net. 
Each U-Net block includes a residual module and two transformer modules, equipped with a temporal attention layer and a spatial attention layer. 
The input condition image  $\textbf{x}$ is processed in two ways. First, it is encoded into a latent space state and then concatenated with the noisy latent sequence $\textbf{z}_t$ for denoising.
Second,  $\textbf{x}$ is used as guidance for the cross-attention mechanism with $\textbf{z}_t$.
This guidance ensures the denoising U-Net generates content aligned with the input identity. Typically, CLIP~\cite{clip} embeddings are used as guidance to provide semantic information for general images.  
Besides, we additionally employ identity guidance, utilizing ArcFace~\cite{deng2019arcface}, a network specifically designed for facial representation, to better preserve facial identity. 
At the same time, the camera pose $\textbf{e}$ is integrated with timestep $\textbf{t} $ and fed into the residual block, ultimately producing frames of multi-view surround images. The fine-tuning objective is defined as follows: 
\begin{equation}
    \mathop{min}\limits_{\theta} \mathrm{E}_{z\sim  \varepsilon(x),t,\epsilon \sim N(0,1)}  ||\epsilon  - \epsilon _{\theta }(\textbf{z}_t;\textbf{t},\textbf{e},\textbf{x})||_2^2
    ,
\end{equation}
 $\theta$ represents the network parameters, $\epsilon$ is the scheduled Gaussian noise added during training, and $\epsilon _{\theta }$ is the predicted noise. 
 The camera pose  \textbf{e} controls the vertical viewpoint of the multi-view images. After the latent diffusion process, the latent state is decoded by the VAE decoder  \textbf{D} to generate the multi-view images. 

 \noindent\textbf{Identity Guidance:}
Previous approaches \cite{liu2023zero, shi2023zero123++, long2024wonder3d, wang2023imagedream} leverage CLIP as a conditional guidance mechanism. However, since CLIP is trained solely for image-text alignment, it lacks the capability to effectively comprehend and represent more sophisticated facial attributes such as identity information and expression variations. To provide better facial information guidance for diffusion models, we leverage ArcFace \cite{deng2019arcface} for improved cross-attention signals. Notably, ArcFace also serves as a powerful classifier; thus, we fine-tune the network on datasets specifically designed for facial expressions and identities to further enhance facial representation.
Through this finetuning process, the ArcFace embedding is capable of distinguishing between different expressions of the same identity while providing consistent representations across different viewing angles.


\noindent\textbf{Two-Stage Diffusion Process:}
While single-stage diffusion models effectively capture global structural information in early denoising steps and refine local details in later stages~\cite{ediff}, such an approach still exhibits limitations in the domain of multi-view human head generation. Specifically, it shows deficiencies in maintaining detailed feature coherence (particularly in eyes and teeth) and produces suboptimal texture clarity. Inspired by DS-VTON~\cite{dsvton}, we introduce a dual-scale framework that explicitly disentangles head structure maintenance and facial texture enhancement.
We first introduce a low-resolution stage to guide the generation process. The proposed process suppresses high-frequency information to guide the diffusion model toward generating multi-view consistent head structures while providing coarse facial landmark priors (e.g., eyes, nose, mouth) for geometric guidance. During training, we downsample ground truth images $\textbf{J}^*$ to $h\times w$ and these images are encoded with the latent state as $\textbf{z}_0$, produced by the VAE encoder \textbf{E}. Then we add noise to the images, predict and remove it using a denoising U-Net, and finally decode the denoised output through the VAE Decoder \textbf{D} to reconstruct low-resolution images $\bar \textbf{J}$. 
Then, a high-resolution diffusion model is introduced for final results generation. In this stage, we extend a residual-based denoising strategy that predicts the residual between high-resolution images and their low-resolution counterparts. This enables the model to focus specifically on texture restoration, building upon the structural alignment achieved in the first stage.
Specifically, we encode ground truth images $\textbf{J}^*$ into the latent space using VAE encoder \textbf{E} and denote them as $\bar \textbf{z}_0$. Meanwhile, we upsample the first-stage results $\bar \textbf{J}$ to $H\times W$, then encode them into the latent space as $\bar \textbf{j}$. Next, we blend $\bar \textbf{j}$ and Gaussian noise $\epsilon$ in a predefined ratio to synthesize the noisy input for denoising. Under this formulation, for each latent noisy sequence $\bar \textbf{z}_t$ the forward and reverse diffusion processes become:
\begin{equation}
    \bar \textbf{z}_t = \sqrt \alpha_t \bar \textbf{z}_{t-1} + \sqrt{1-\alpha_t}(\xi_1 \cdot \epsilon + \xi_2 \cdot \bar \textbf{j}),
\end{equation}
\begin{equation}
    \bar \textbf{z}_{t-1} = \frac{1}{\sqrt{\alpha_t}}(\bar \textbf{z}_t-\frac{1-\alpha_t}{\sqrt{1-\bar \alpha_t}}\bar\epsilon_{\bar\theta}(\bar \textbf{z}_t, \textbf{t}, \textbf{e}, \textbf{x}))+\sigma_t\textbf{z},
\end{equation}
where $\xi_1$ and $\xi_2$ represent the noise weights, $\bar \epsilon_{\bar \theta}$ represent the predicted
noise in this stage, other notations follow DDPM\cite{ho2020denoising}. Through this residual-guided denoising process, we obtain texture-enhanced high-resolution multi-view images $\textbf{J}$, providing high-quality inputs for downstream reconstruction tasks.

\noindent\textbf{Objective Functions:} To improve generalization, we add identity loss to our loss function. The identity loss computes the cosine distance between the latent features of \textbf{J} and the ground truth \textbf{J}$^{*} \in \mathrm {R}^{3 \times H\times W }$ extracted from ArcFace network. It is defined as:
\begin{equation}
    \mathcal{L}_{ID}(\textbf{J}, \textbf{J}^{*}) = 1 -\frac{1}{N_s}\sum_{i}^{N_s}\frac{j_{i} \cdot j_{i}^*}{||j_i||||j_i^*||},
\end{equation}
where $j_{i}, j_{i}^*$ represent features of \textbf{J} and \textbf{J}$^{*}$. We finally employ identical loss functions for both stages, including MSE loss, perceptual loss~\cite{lpips}, and ID loss. The overall loss function is defined as:

\begin{equation}
    \mathcal{L_M}(\textbf{J}, \textbf{J}^{*}) = \mathcal{L}_{MSE} + \lambda_p \mathcal{L}_{perc} +
    \lambda_i \mathcal{L}_{ID}
    ,
\end{equation}
where $\lambda_p$ and $\lambda_i$ are loss weights. 

\begin{figure*}[t]
    \centering
    \includegraphics[width=0.95\linewidth]{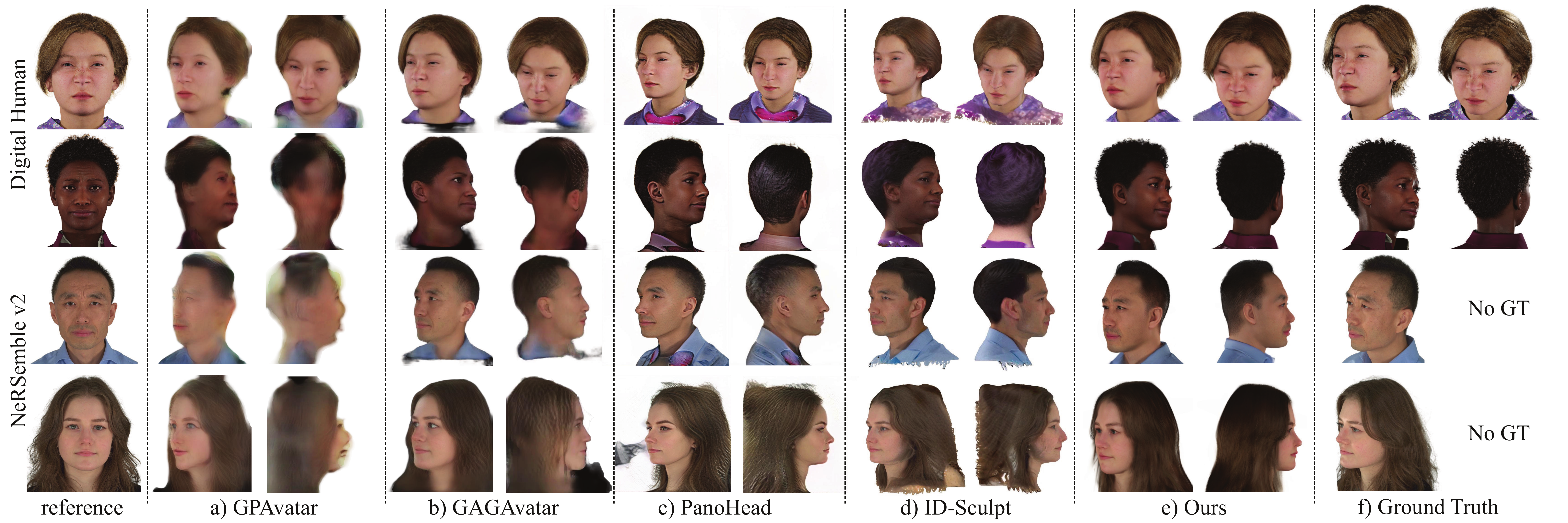}
    \caption{Qualitative comparison. Visualization results demonstrate that our method significantly outperforms existing approaches in capturing fine-grained details, including head geometry, facial feature consistency, expression texture, and gaze direction.}
    \label{fig:whole_head}
\end{figure*}

\begin{table*}[]
\centering
\begin{tabular}{c|cccccc}
\hline
Dataset                        & Method     & LPIPS$\downarrow$        & PSNR$\uparrow$            & CSIM$\uparrow$           & DreamSim$\downarrow$     & User Study$\uparrow$  \\ \hline
\multirow{5}{*}{Digital Human} & GPAvatar   & 0.7452 ± 0.0050          & 10.1993 ± 0.0773          & 0.6461 ± 0.0127          & 0.3424 ± 0.0092          & 0.0             \\
                               & GAGAvatar  & 0.2960 ± 0.0081          & 13.2312 ± 0.1907          & 0.8808 ± 0.0078          & 0.1611 ± 0.0050          & 0.0             \\
                               & PanoHead   & 0.3230 ± 0.0101          & 13.0568 ± 0.5874          & 0.8603 ± 0.0132          & 0.1453 ± 0.0080          & 0.0650           \\
                               & ID-Sculpt  & 0.3145 ± 0.0111          & 12.5297 ± 0.5861          & 0.8483 ± 0.0142          & 0.2080 ± 0.0159          & 0.0215          \\
                               & Ours       & \textbf{0.2315 ± 0.0054} & \textbf{20.4287 ± 0.2082} & \textbf{0.9189 ± 0.0062} & \textbf{0.0804 ± 0.0031} & \textbf{0.9135} \\ \hline
\multirow{5}{*}{NeRSemble v2}  & GPAvatar   & 0.7081 ± 0.0024          & 10.8568 ± 0.1521          & 0.5336 ± 0.0130          & 0.2740 ± 0.0056          & 0.0             \\
                               & GAGAvatar & 0.3329 ± 0.0090          & 13.2822 ± 0.3308          & 0.8424 ± 0.0132          & 0.1370 ± 0.0040          & 0.0215          \\
                               & PanoHead   & 0.3649 ± 0.0120          & 13.3809 ± 0.3605          & 0.8379 ± 0.0118          & 0.1128 ± 0.0072          & 0.0655          \\
                               & ID-Sculpt  & 0.3284 ± 0.0085          & 13.7613 ± 0.4523          & 0.8293 ± 0.0090          & 0.1587 ± 0.0105          & 0.0             \\
                               & Ours       & \textbf{0.3147 ± 0.0099} & \textbf{16.0120 ± 0.3365} & \textbf{0.8519 ± 0.0115} & \textbf{0.1043 ± 0.0033} & \textbf{0.9130} \\ \hline
\end{tabular}
\caption{Quantitative evaluation of single image to 3D head. We conduct a comprehensive evaluation of our method against 4 state-of-the-art 3D head generation approaches across two distinct datasets, employing five diverse metrics. Experimental results demonstrate that our approach consistently outperforms all competing methods, achieving superior performance across all evaluation criteria.}
\label{tab:whole_head}
\end{table*}

\subsection{3D Gaussian Avatar Reconstruction}\label{ff}
 We reconstruct a 3D Gaussian avatar from the multi-view images $\textbf{J}$ and their corresponding camera positions $\pi$. Although existing feed-forward Gaussian point cloud generation methods \cite{lgm, gslrm} have achieved remarkable success in object reconstruction from sparse-view inputs, they still struggle to faithfully recover facial textures and fine-grained details in human head reconstruction. To address this, we utilize a feed-forward Gaussian reconstruction approach \cite{lgm} for enhanced reconstruction. The entire process is illustrated in the lower segment of Fig.~\ref{fig:pipeline}. 

\noindent\textbf{Architecture:} The method proposes a 3D U-Net for Gaussian point cloud generation. Each block in this U-Net comprises a residual layer and a 3D self-attention layer~(two spatial dimensions and one dimension across input images). Given 4 input images $\hat{\textbf{J}}$ and corresponding camera parameters $\hat{\pi}$, the method first calculates each camera’s world coordinate origin $o_n \in \mathrm{R} ^ 3$ and ray direction $d_n \in \mathrm{R} ^3$ for each pixel, and then the method concatenates $\hat{\textbf{J}}$, the corresponding Plücker ray embedding $o_n \times d_n$ and the ray direction $d_n$ in the last channel as the 3D U-Net's inputs. Finally, the U-Net outputs a feature map shaped as [4, $H'$, $W'$, 14], and by flattening the first three channels, a Gaussian point cloud $\textbf{P}$ is generated.

\noindent\textbf{Single Subject Optimization:}
Building upon the 3D U-Net architecture, we fine-tune the network on our digital human head dataset to optimize performance. However, empirical evaluation indicated that the one-step generation results exhibit deficiencies in both facial texture details and multi-view consistency.
To address this issue, we leverage the availability of multi-view images from our upstream pipeline to perform individualized optimization for each subject.
Specifically, we choose multi-view input configuration comprising 4 equidistant perspectives $\hat{\textbf{J}}$ (frontal, right, posterior, and left views) to generate a Gaussian point cloud representation \textbf{P}. Subsequently, we utilize the camera parameters associated with \textbf{J} to render synthesized images $\textbf{J}^{**}\in \mathrm{R}^{N_s \times 3\times H\times W} $, which are then iteratively optimized against the corresponding multi-view reference images from ID-TS diffusion process. The final loss function aligns with the ID-TS model, comprising MSE loss, perceptual loss, and ID loss.
With this optimization strategy, we significantly enhance the quality of 3D reconstruction for each subject, achieving accurate facial details and multi-view consistency.

\section{Experiments}

\subsection{Experimental Settings}
During the multi-view stage, our model generates $N_S=16$ multi-view head images with an initial learning rate of 1e-6 and a cosine annealing scheduler, setting the loss function coefficients $\lambda_p=\lambda_i=1.0$; the low-resolution stage operates at $h\times w=256\times256$ resolution and trains for ~20h on 8$\times$A6000 GPUs, while the high-resolution stage uses $H\times W=512\times512$ resolution with a  mix ratio $\xi_1=0.4, \xi_2=0.6$ of low-resolution inputs and noise(mixing ratio experiment is shown in supplementary material), training for ~4 days on 8$\times$A6000 GPUs; for the Gaussian reconstruction network, we set the learning rate to 1e-5 and the loss function coefficients $\lambda_p=\lambda_i=1.0$, outputting a $4\times H'\times W'\times14 = 4\times128\times128\times14$ tensor after ~2 days of training on 8$\times$A6000 GPUs. 
We also perform per-subject 3D U-Net optimization (LR=1e-6) on a single A6000 GPU; it takes ~400 s in total, with seed 42 for reproducibility.

\noindent\textbf{Datasets.}
Our experiments are conducted on two distinct datasets for comprehensive evaluation. The first dataset is the NeRSemble v2~\cite{nersemble}, where we select 14 subjects, each featuring 2 distinct expressions.
The second dataset is our newly constructed digital human head dataset, comprising previously unseen faces and expressions, where we select 10 subjects, each with 5 distinct expressions. 
For quantitative evaluation, we randomly sample 16 multi-view facial images, while for qualitative assessment, the test viewpoints are not constrained to frontal faces. As for ablative analyses, we only use digital human head  because of the viewpoint restriction.

\noindent\textbf{Baselines.}
For the evaluation of the entire pipeline, from a single image to a 3D head representation, we selected single-image-to-3D-head methods that provide publicly available implementations, including GPAvatar~\cite{gpavatar}(ICLR 2024), GAGAvatar~\cite{gagavatar}(NeurlPS 2024), PanoHead~\cite{panohead}(CVPR 2023), and ID-Sculpt~\cite{portrait3d}(AAAI 2025). 

\begin{figure*}[t]
    \centering
    \includegraphics[width=0.9\linewidth]{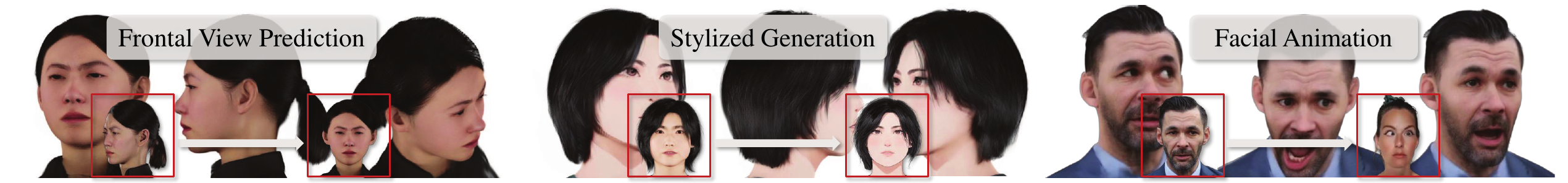}
    \caption{Applications for downstream tasks. We evaluate our method on three downstream tasks: frontal view prediction from profile views, stylized head generation, and facial animation, demonstrating its versatility across diverse application scenarios.}
    \label{fig:app}
\end{figure*}

\begin{table*}[]
\centering
\begin{tabular}{c|ccccc}
\hline
Category                                                                                & Ablation  & LPIPS$\downarrow$        & PSNR$\uparrow$            & CSIM$\uparrow$           & DreamSim$\downarrow$     \\ \hline
\multirow{2}{*}{\begin{tabular}[c]{@{}c@{}}Low-Resolution Stage\\ 256x256\end{tabular}} & Fine-tuned & 0.2816 ± 0.0080          & 18.6968 ± 0.3564          & 0.8860 ± 0.0087          & 0.1217 ± 0.0068          \\
                                                                                        & +ID       & \textbf{0.2770 ± 0.0075} & \textbf{18.9627 ± 0.2307} & \textbf{0.8961 ± 0.0100} & \textbf{0.1106 ± 0.0063} \\ \hline
\multirow{4}{*}{\begin{tabular}[c]{@{}c@{}}ID-TS Diffusion\\ 512x512\end{tabular}}      & Fine-tuned & 0.2446 ± 0.0063            & 17.5717 ± 0.4647          & 0.8799 ± 0.0098          & 0.0938 ± 0.0064          \\
                                                                                        & +ID       & 0.2414 ± 0.0036          & 18.2606 ± 0.1940          & 0.9039 ± 0.0062          & 0.0870 ± 0.0030          \\
                                                                                        & +TS       & 0.2092 ± 0.0040          & 19.1919 ± 0.4311          & 0.9164 ± 0.0072          & 0.0603 ± 0.0026          \\
                                                                                        & +ID+TS    & \textbf{0.2028 ± 0.0034} & \textbf{19.9974 ± 0.2503} & \textbf{0.9261 ± 0.0071} & \textbf{0.0585 ± 0.0022} \\ \hline
\multirow{3}{*}{Gaussian Head Generation}                                               & O         & 0.2980 ± 0.0044          & 15.6821 ± 0.0846          & 0.8578 ± 0.0067          & 0.1413 ± 0.0032          \\
                                                                                        & O+H       & 0.2556 ± 0.0048          & 18.3844 ± 0.1326          & 0.9005 ± 0.0039          & 0.1036 ± 0.0039          \\
                                                                                        & O+H+S     & \textbf{0.2315 ± 0.0054} & \textbf{20.4287 ± 0.2082} & \textbf{0.9189 ± 0.0062} & \textbf{0.0804 ± 0.0031} \\ \hline
\end{tabular}
\caption{Quantitative ablation results for our three core contributions. Both ID guidance and the two-stage diffusion process show higher scores, and per-subject optimization presents better results among different reconstruction strategies. }
\label{tab:ablation}
\end{table*}

\noindent\textbf{Evaluation Metric.}
We compared each generated frame with its corresponding ground truth frame using LPIPS~\cite{lpips}, PSNR, DreamSim~\cite{dreamsim}, and cosine similarity of identity embeddings~(CSIM)~\cite{deng2019arcface}. These metrics collectively assess both pixel-level accuracy and identity preservation. Additionally, we conducted a questionnaire-based user study, selecting four test cases from each comparison category and asking participants to choose the best generation method. We then analyzed the distribution of preferences based on the proportion of responses.

\subsection{Single Image to 3D head Evaluation} \label{whole}
We evaluated the whole pipeline from single head image input to 3D full-head representation. We observe that varying camera coordinate systems across different methods lead to inconsistent head sizes and positions in the sampled multi-view images. To mitigate this, we align the head keypoints of images generated by different methods, minimizing potential testing bias. Quantitative results are shown in Tab.~\ref{tab:whole_head}, and qualitative comparison is presented in Fig.~\ref{fig:whole_head}. Among these methods, GPAvatar struggles with large-angle profile generation, GAGAvatar exhibits errors in gaze direction, PanoHead lacks fidelity in clothing and expression preservation, while ID-Sculpt suffers from inaccurate shape estimation. Both experiments demonstrate that our method achieves the best results in generating realistic head shapes and fine-grained facial texture details. Additional experiments and dynamic video comparisons are provided in the supplementary material.

\subsection{Ablative Analyses} \label{aa}
\begin{figure}
    \centering
    \includegraphics[width=0.80\linewidth]{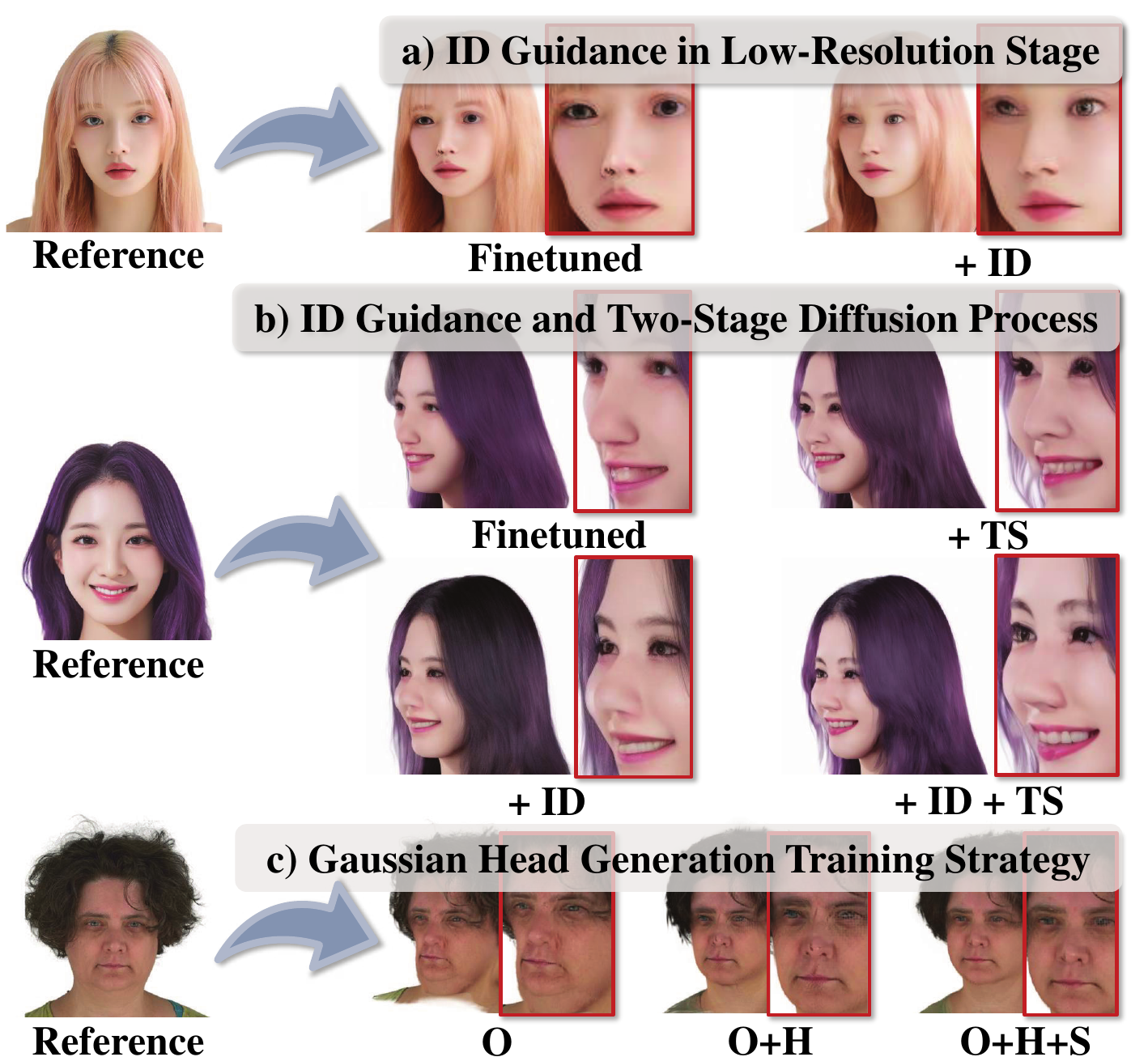}
    \caption{Visual results for ablation study. Visualization demonstrates that our innovation significantly enhances the quality of multi-view and 3D head model reconstruction.}
    \label{fig:ablation}
\end{figure}

For ablative analyses, we first evaluate the effect of ID guidance at the low-resolution stage. Then, we examine the impact of incorporating ID guidance in ID-TS diffusion and using a two-stage diffusion process. Finally, we test different 3D point cloud generation strategies, and we define that ``O'' represents training with 3D object dataset, ``H'' represents fine-tuning with 3D digital human dataset, ``S'' represents optimizing in single subject. Quantitative experiments are performed solely on the digital head dataset, as shown in Tab.~\ref{tab:ablation}. And visual comparison results are presented in Fig.~\ref{fig:ablation}. Experimental results demonstrate that identity guidance preserves facial fidelity, while the two-stage diffusion process enhances texture details. Our single-head optimization reduces Gaussian splatting artifacts, improving 3D reconstruction quality.

\subsection{Applications} \label{app}
We demonstrate three representative downstream applications enabled by our 3D reconstruction framework, with qualitative results visualized in Fig.~\ref{fig:app}.

\noindent\textbf{Frontal View Prediction.}
Frontal view prediction can facilitate applications such as suspect facial reconstruction and access control enhancement in face recognition. Our approach employs randomly sampled viewpoint conditions during multi-view training process, enabling the model to generate high-quality frontal face images even from extreme profile views.

\noindent\textbf{Stylized Generation.}
Stylized 3D head generation enables the creation of virtual 3D avatars and facilitates applications in film production. We propose a framework that constructs high-fidelity 3D anime-style characters by first translating 2D images into stylized artwork and then integrating the processed head into our model for 3D reconstruction.

\noindent\textbf{Facial Animation.}
The construction of 4D head avatars enables applications in virtual video conferencing, virtual streaming, and related domains. We achieve 4D head avatar synthesis by sampling video outputs from our 3D head model and leveraging a 2D facial expression transfer framework~\cite{liveportrait}.

\section{Conclusion}

Although our method achieves promising results, limitations remain in image resolution due to hardware constraints and in generalization to unseen accessories, which we aim to address through improved hardware and dataset expansion.

\section*{Acknowledgments}
This research is supported, in part, by the National Natural Science Foundation of China (Grant No. 62302295), the Shanghai Municipal Science and Technology Major Project, China (Grant No. 2021SHZDZX0102), the Startup Fund for Young Faculty at SJTU (Grant No. 22X010503821) and the foundation of Key Laboratory of Artificial Intelligence, Ministry of Education, P.R. China.

\bibliography{aaai2026}

\end{document}